# Opportunistic Self Organizing Migrating Algorithm for Real-Time Dynamic Traveling Salesman Problem

Shubham Dokania*, Sunyam Bagga†, and Rohit Sharma‡
*shubham.k.dokania@gmail.com, †sunyambagga@gmail.com, ‡deltechrs2119@gmail.com
†‡Department of Computer Science, *Department of Applied Mathematics
Delhi Technological University
New Delhi, India

*Abstract*—Self Organizing Migrating Algorithm (SOMA) is a meta-heuristic algorithm based on the self-organizing behavior of individuals in a simulated social environment. SOMA performs iterative computations on a population of potential solutions in the given search space to obtain an optimal solution. In this paper, an Opportunistic Self Organizing Migrating Algorithm (OSOMA) has been proposed that introduces a novel strategy to generate perturbations effectively. This strategy allows the individual to span across more possible solutions and thus, is able to produce better solutions. A comprehensive analysis of OSOMA on multi-dimensional unconstrained benchmark test functions is performed. OSOMA is then applied to solve real-time Dynamic Traveling Salesman Problem (DTSP). The problem of real-time DTSP has been stipulated and simulated using real-time data from Google Maps with a varying cost-metric between any two cities. Although DTSP is a very common and intuitive model in the real world, its presence in literature is still very limited. OSOMA performs exceptionally well on the problems mentioned above. To substantiate this claim, the performance of OSOMA is compared with SOMA, Differential Evolution and Particle Swarm Optimization.

## I. INTRODUCTION

Optimization is a mathematical process to obtain the best solution out of a pool of certain possible solutions. Over the years, several algorithms inspired from natural phenomena have been proposed to efficiently solve various optimization problems. Evolutionary Algorithms are classified as meta-heuristic search algorithms, where possible solution elements span the n-dimensional search space to find the global optimum solution. These algorithms perform efficiently on various real-world problems. Most of these algorithms involve the creation of new solutions and discard those which fail to produce suitable results. Bacterial Foraging Optimization Algorithm (BFOA) [1] [2] explains a variety of bacterial swarming and social foraging behavior and dictates how foraging should proceed in E. coli. Ant Colony Optimization (ACO) [3] deals with the complex social behavior of ants which provides models for solving difficult combinatorial optimization problems. Particle Swarm Optimization (PSO) is a swarm intelligence algorithm based on the social behavior of a group of individuals such as flocking behavior of birds, school of fishes [4]. These individuals, called particles, move through an n-dimensional search space and share their information in order to find the global optimum. Differential Evolution (DE) is a stochastic, population based meta-heuristic algorithm in which random candidate solutions are generated [5]. SOMA is another evolutionary algorithm which was proposed by Zelinka et al. in 2000 [6]. It follows a cooperative-competitive behavior which differs from other algorithms in its approach to find new solutions. There are various strategies used for SOMA such as All-to-One and All-to-All. In All-to-One approach, the leader acts as a reference point for all other active individuals. In All-to-All approach, each individual moves towards all other agents which may lead to higher computational complexity [7]. All-to-One approach has been considered in this paper.

These nature-inspired algorithms do not yield an exact solution for an optimization problem but a close-to-optimal solution can be achieved. So, there is always scope for improvement in the efficiency of such algorithms. Kusum Deep et al. [8] proposed a binary coded Self Organizing Migrating Genetic Algorithm which is a hybridization of simple binary coded GA with real coded SOMA. Dipti et al. have developed a hybridization of SOMA with Quadratic Approximation crossover operator and Log Logistic mutation operator to maintain the diversity of the population and increase the search capability [9].

Traveling Salesman Problem (TSP) is a well known NP-hard problem in combinatorial optimization which was first studied in the 1930s by Karl Menger in Vienna and Harvard. Dynamic Traveling Salesman Problem is an extension of the traditional TSP which involves dynamic-cost allocation between the cities. This may include, but is not limited to, inclusion and exclusion of certain cities at any given time. It was proposed by Psaraftis [10] in 1988. DTSP is widely applicable in real-time scenarios than TSP but is arguably more difficult to solve. Usually, the algorithms used to solve Static TSP turn out to be inefficient for DTSP. In this paper, a real-time DTSP has been stipulated by using real time traffic data feed from Google Maps [11].

## II. SELF ORGANIZING MIGRATING ALGORITHM

SOMA is based on the cooperation among a population of individuals (agents) where a stochastic search technique is employed for achieving global optimization. It is classified as an evolutionary algorithm; however, no new individuals are created during the search process. The robustness of this algorithm is evident from its fast convergence to the global extreme.

Each iteration in SOMA is called a migration loop, in which each individual locates the best position by competing against other individuals. The individual with the best fitness value becomes the leader and active individuals travel in the search space by following the leader. They exchange information (cooperate) to update their leader after each migration loop. Thus, individuals in SOMA follow cooperative-competitive behavior. In each migration loop, all active individuals move towards the leader as shown in Fig. 1. It is a contour plot of how the individuals traverse the search space in a particular migration loop. The black solid points on the path of an individual represent the set of possible positions available in SOMA. The red points represent the current position of the individuals; the leaders position is represented by the green point which influences the path of other individuals. The global optimum is represented by the blue cross. The pseudo code for SOMA is shown in Algorithm 1. Movement of individuals is governed by eq. (1) as shown:

$$x_{i+1} = x_i + (x_L - x_i) * \phi * L \quad (1)$$

$$\phi = \begin{cases} 1, & \gamma < PR \\ 0, & otherwise \end{cases} \quad (2)$$

where, $x_i$ is the current position of the individual,
$x_{i+1}$ is the next position of the current individual,
$x_L$ is the position of the leader,
$L$ is the Path Value or displacement from the initial position for the current individual,
$\phi$ is the perturbation vector,
$PR$ is the probability of perturbation,
$\gamma \in U(0,1)$ is a random number from a uniform distribution.

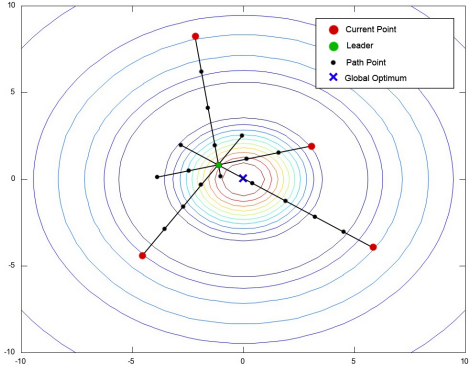

Fig. 1: Movement of individuals in a migration loop in SOMA

**Algorithm 1** Self Organizing Migrating Algorithm
1: **procedure** START
2:     Build initial population $P$.
3:     Initialize all parameters.
4:     **while** termination criteria is not met **do**
5:         **for** each individual $x_i$ in $P$ **do**
6:             Evaluate objective $Z$ for $x_i$.
7:         **end for**
8:         Choose leader using $Z$.
9:         **for** each individual $x_i$ in $P$ **do**
10:            Generate $\phi$ using (2).
11:            Update position using (1).
12:         **end for**
13:     **end while**
14: **end procedure**

## III. OPPORTUNISTIC SELF ORGANIZING MIGRATING ALGORITHM

SOMA is a relatively new stochastic optimization algorithm based on the social behavior of cooperating individuals. Active individuals in SOMA move towards the leader based on the value of perturbation vector. In each dimension, they can either move a full step-size or not move at all. In our modified approach, an opportunistic strategy for generating the perturbation vector is proposed. The uniqueness of this algorithm is that it enhances the capability of an individual by providing it with a higher number of choices than SOMA. This allows OSOMA to exploit the search space and thus, yield better solutions. The individuals in OSOMA have their motion governed by a strategy which prohibits them to attain a stationary state. This strategy is quantitatively expressed as shown in eq. (3)

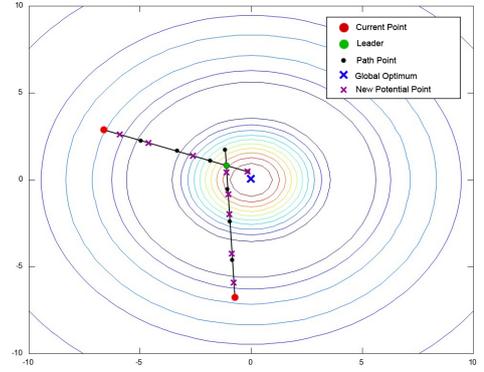

Fig. 2: Movement of individuals in a migration loop in OSOMA

$$x_{i+1} = x_i + (x_L - x_i) * \phi * L \quad (3)$$

$$\phi = \begin{cases} 1, & \gamma < PR \\ \lambda/D_n, & otherwise \end{cases} \quad (4)$$

where, $x_i$ is the current position of the individual,
$x_{i+1}$ is the next position of the current individual,
$x_L$ is the position of the leader,
$L$ is the Path Value or displacement from the initial position for the current individual,
$\phi$ is the perturbation vector,
$PR$ is the probability of perturbation,
$\gamma \in U(0,1)$ is a random number from a uniform distribution,
$\lambda \in U(0.60, 0.85)$ is a random number from uniform distribution, where limits have been obtained by experimental analysis,
$D_n$ is the dimensionality of the problem.
The expression $\lambda/D_n$ returns a small positive value which allows the individual to search through a higher number of potential solutions. $D_n$ is the dimensionality of the objective function which stabilizes the perturbation value. The pseudo code for OSOMA is shown in Algorithm 2.

Fig. 2 is an illustration of how the individuals traverse the search space in a particular migration loop. The solid points on the path of an individual represent the set of possible positions available in SOMA. OSOMA follows the migration loop as shown in eq. (3). The set of crosses represents the new potential positions generated in OSOMA which were not available to individuals in SOMA. This set represents a particular case and these crosses can be present anywhere along the direction

**Algorithm 2** Opportunistic Self Organizing Migrating Algorithm

1: **procedure** START
2:     Build initial population $P$.
3:     Initialize all parameters.
4:     **while** termination criteria is not met **do**
5:         **for** each individual $x_i$ in $P$ **do**
6:             Evaluate objective $Z$ for $x_i$.
7:         **end for**
8:         Choose leader using $Z$.
9:         **for** each individual $x_i$ in $P$ **do**
10:            Generate $\phi$ using (4).
11:            Update position using (3).
12:         **end for**
13:     **end while**
14: **end procedure**

of traversal. Hence, the proposed algorithm allows individuals to effectively exploit the given search space. The number of dimensions plays an important role in this approach. When the number of dimensions increases, the average perturbation value returned decreases. Hence, performance of OSOMA approaches that of SOMA at higher dimensions.

## IV. REAL-TIME DYNAMIC TRAVELING SALESMAN PROBLEM

Traveling Salesman Problem is an intensively studied problem in graph theory, and is used as a benchmark for various optimization problems. The problem statement for the Traveling Salesman Problem is as follows: "A traveling salesman has to find the shortest tour to visit $N$ number of cities, with each city being visited exactly once except the first city, which is visited twice.". It belongs to the class of NP-Hard problems and no effective solution method is yet known for the general case.

Real Time Dynamic Traveling Salesman Problem is an extension of the Traveling Salesman Problem where the cost matrix is time-varying. The cost of traveling from city $i$ to city $j$ changes with time or with an addition of a new city in the network. The expression for the dynamic cost matrix is given in eq. (5).

$$D(t) = [d_{ij}(t)] \quad (5)$$

where,
$D(t)$ represents the time dependent cost matrix,
$d_{ij}(t)$ represents the cost to travel from city $i$ to city $j$ as a function of time $t$.

In real-world, external factors such as traffic may lead to a change in the traveling time of the salesman. Hence, the cost matrix is dynamically updated resulting in an even more complex problem than the Traveling Salesman Problem. The dynamic nature of this problem requires a continuous evaluation due to the updates in the cost matrix. The path to be followed by the salesman is dynamic in nature i.e. bound to change continuously.

OSOMA has been applied on real-time DTSP using a special type of encoding proposed by Hadia et al. [12]. They introduced a new concept of Swap Operations to solve the city routing problem. This concept has been used in the application of OSOMA on real-time DTSP. Their concept of swap is presented with the help of an example:

Let $M$ represent a possible tour, then $N = M \oplus MO(i, j)$ provides a new tour generated on application of operator $MO(i, j)$. For example:
Let $M = (3, 4, 5, 6, 8)$.
Then, $M' = M \oplus MO(2, 3) = (3, 4, 5, 6, 8) \oplus MO(2, 3) = (3, 5, 4, 6, 8)$
A Swap Sequence can be defined as the sequence of Swap Operators.
Let $A = (5, 6, 7, 8, 9)$ and $B = (6, 7, 5, 9, 8)$.
As $A(1) = B(3) = 5$, so first swap operation will be $MO(1, 3)$; $B' = B \oplus MO(1, 3) = (5, 7, 6, 9, 8)$
Now, $A(2) = B(3) = 6$, so second swap operation will be $MO(2, 3)$; $B'' = B' \oplus MO(2, 3) = (5, 6, 7, 9, 8)$.
Now, $A(4) = B(5) = 8$, so third swap operation will be $MO(4, 5)$; $B''' = B'' \oplus MO(4, 5) = (5, 6, 7, 8, 9)$
Hence, the final Swap Sequence $MM = A - B = [MO(1, 3), MO(2, 3), MO(4, 5)]$.
Using the above encoding method, the migration loop update equation of OSOMA (3) can be rewritten as follows:

$$x_{i+1} = x_i \oplus (x_L - x_i) * \phi * L \quad (6)$$

Where,
$\oplus$ and $-$ are the operators as shown in the above example, and the rest of parameters are same as in (3). The value obtained from $\phi * L$ is used to model the probability which picks the elements from $(x_L - x_i)$ to participate in the sequence update procedure, hence making $\phi * L$ a probabilistic decision factor.

## V. RESULTS AND DISCUSSIONS

All the evaluations were performed in python (2.7.6) using Scipy and Numpy [13] frameworks for scientific computations and Matplotlib [14] package for graphical representation of the results. This section has been divided into two subsections: V-A discusses the results obtained on standard benchmark test functions, V-B presents the results obtained on applying OSOMA to real-time DTSP.

### A. Standard benchmark Test Functions

Table I [15] lists the test functions under consideration. It should be noted that all test functions have a global minima of 0.0 except Easom which has a global minimum of -1.0. Tables II and III show the average values obtained for these functions by OSOMA, SOMA, PSO and DE. The results have been obtained for dimensions ($d = 2, 5$). Although OSOMA works on higher dimensions as well, only lower dimensions have been taken into account due to the nature of real-time DTSP under consideration.

Table II shows the results of considered algorithms on 11 two-dimensional test functions. SOMA, DE and PSO fail to converge to global minima in many cases while OSOMA attains the optimum value each time. In f2 test function, SOMA fails to converge to global optimum (attaining $4.365e - 16$) while OSOMA achieves 0.0. Similarly, all the other algorithms are outperformed by OSOMA. In f3, DE, PSO and OSOMA get stuck at the same local minima with value $1.183e - 30$ as shown in Table III. Here, SOMA is also stuck at a local minimum but with a worse value. This is because f3 is a relatively complex multimodal minimization test function.
The graphs in Fig. 3 show the objective value attained in initial iterations by PSO, DE, SOMA and OSOMA on f1 for dimensions $d = 2, 5$. It can be seen that the convergence rate of DE is the worst among considered algorithms. OSOMA reaches the optimal value of 0.0 each time and at a considerably faster rate than the other algorithms. Hence, the accuracy of OSOMA is substantially higher than that of PSO, DE and SOMA as seen in Tables II and III. This can be attributed to the opportunistic nature of perturbations generated in OSOMA which enhances the ability of an individual to explore the search space efficiently. It is evident from the discussion above that OSOMA performs well as compared to the other algorithms.

TABLE I: Test Functions

| Name | Problem Function | Interval |
|---|---|---|
| Sphere (f1) | $f(x) = \sum_{i=1}^{d} x_i^2$ | $[-5.12, 5.12]$ |
| Ackley (f2) | $f(x) = -20\exp(-0.2\sqrt{\frac{1}{a}\sum_{i=1}^{d} x_i^2}) - \exp(\frac{1}{a}\sum_{i=1}^{d}\cos(cx_i)) + 20 + e$ | $[-32, 32]$ |
| Qing (f3) | $f(x) = \sum_{i=1}^{d}(x_i^2 - i)^2$ | $[-500, 500]$ |
| $3^{rd}$ De Jong (f4) | $f(x) = \sum_{i=1}^{d} |x_i|$ | $[-2.048, 2.048]$ |
| $4^{th}$ De Jong (f5) | $f(x) = \sum_{i=1}^{d} ix_i^4$ | $[-1.28, 1.28]$ |
| Rosenbrock (f6) | $f(x) = \sum_{i=1}^{d}[100(x_{i+1} - x_i^2)^2 + (x_i - 1)^2]$ | $[-100, 100]$ |
| Schwefel (f7) | $f(x) = (\sum_{i=1}^{d} x_i^2)^\pi$ | $[-100, 100]$ |
| Booth (f8) | $f(x) = (x + 2y - 7)^2 + (2x + y - 5)^2$ | $[-5, 5]$ |
| Matyas (f9) | $f(x) = 0.26(x_1^2 + x_2^2) - 0.48x_1 x_2$ | $[-10, 10]$ |
| Easom (f10) | $f(x) = -\cos(x_1)\cos(x_2)\exp(-(x_1 - \pi)^2 - (x_2 - \pi)^2)$ | $[-100, 100]$ |
| Bohachevsky (f11) | $f(x) = x_1^2 + 2x_2^2 - 0.3\cos(3\pi x_1) - 0.4\cos(4\pi x_2) + 0.7$ | $[-100, 100]$ |

TABLE II: Objective Function Value of DE, PSO, SOMA, OSOMA for $D_n = 2$

| Function | DE | PSO | SOMA | OSOMA |
|---|---|---|---|---|
| f1 | 9.355e-155 | 6.738e-146 | 4.409e-144 | 0.0 |
| f2 | 0.0 | 0.0 | 4.365e-16 | 0.0 |
| f3 | 0.0 | 1.972e-31 | 1.131e-270 | 0.0 |
| f4 | 3.077e-77 | 6.182e-59 | 8.073e-43 | 0.0 |
| f5 | 9.815e-298 | 1.747e-235 | 1.131e-270 | 0.0 |
| f6 | 0.0 | 1.573e-10 | 0.0 | 0.0 |
| f7 | 1.196e-271 | 1.317e-205 | 2.854e-270 | 0.0 |
| f8 | 0.0 | 0.0 | 0.0 | 0.0 |
| f9 | 6.187e-66 | 3.182e-120 | 2.162e-144 | 0.0 |
| f10 | 0.0 | 0.0 | 0.0 | 0.0 |
| f11 | -8.11e-05 | -1.0 | -1.0 | -1.0 |

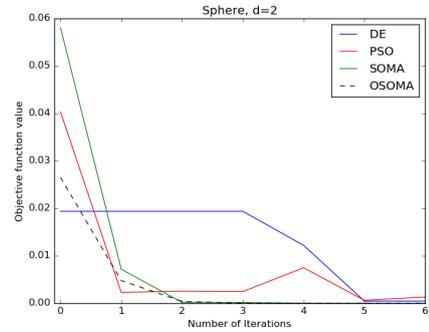

(a)

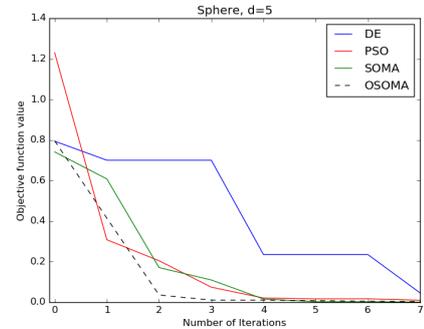

(b)

Fig. 3: Objective function value vs. Number of iterations for $f1$ in: (a) $D_n = 2$, (b) $D_n = 5$

TABLE III: Objective Function Value of DE, PSO, SOMA, OSOMA for $D_n = 5$

| Function | DE | PSO | SOMA | OSOMA |
|---|---|---|---|---|
| f1 | 2.654e-86 | 5.648e-86 | 3.031e-77 | 0.0 |
| f2 | 0.0 | 7.395e-11 | 6.579e-12 | 3.552e-15 |
| f3 | 1.183e-30 | 1.183e-30 | 8.221e-25 | 1.183e-30 |
| f4 | 7.821e-57 | 1.867e-43 | 1.326e-08 | 2.436e-60 |
| f5 | 1.782e-136 | 3.778e-162 | 4.495e-163 | 0.0 |
| f6 | 5.784e-04 | 7.858e-02 | 1.693e-06 | 2.288e-04 |
| f7 | 7.809e-196 | 1.081e-148 | 3.396e-103 | 0.0 |

### B. Real-Time Dynamic Traveling Salesman Problem

Dynamic Traveling Salesman Problem has been studied extensively in the past. A real-time DTSP deals with a dynamic cost matrix depending on conditions such as: Addition of new cities to the tour, Change in the 'cost' to travel from one city to another etc. The comprehensive discussion in this section covers the above-mentioned conditions. In the Initial Case, the graph is initialised with 11 cities as shown in Fig. 4a. In Case 1, a new city is added to the graph while the salesman has partially completed his tour. Case 2 considers the possibility of a change in cost incurred to travel from one location to another.

*1) Initial Case:* The initial case is analogous to the Static Traveling Salesman Problem. OSOMA, SOMA, DE and PSO

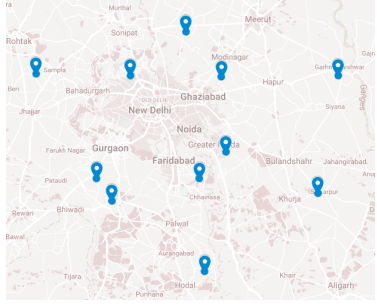

(a) Initial set of 11 cities.

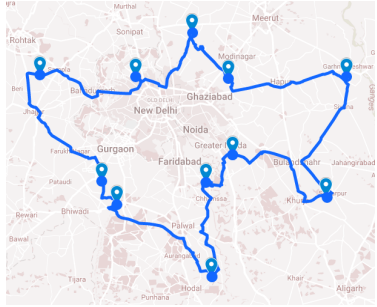

(b) Optimal Tour for the 11 cities.

Fig. 4: Maps depicting initial set of 11 cities and its optimal path

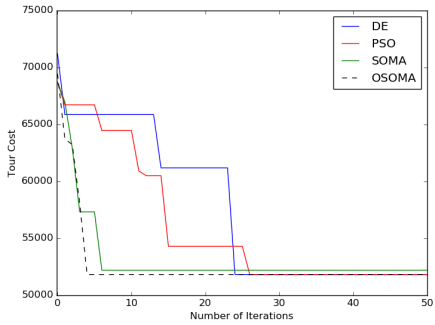

Fig. 5: Tour cost vs. Number of iterations for initial set of 11 cities (for 50 iterations).

TABLE IV: Tour Cost of DE, PSO, SOMA, OSOMA as a function of Number of Iterations

| Iterations | SOMA | DE | PSO | OSOMA |
|---|---|---|---|---|
| 0 | 68701 | 71224 | 68725 | 69459 |
| 5 | 57332 | 65869 | 66728 | 51837 |
| 10 | 52214 | 65869 | 64473 | 51837 |
| 15 | 52214 | 61196 | 54307 | 51837 |
| 20 | 52214 | 61196 | 54307 | 51837 |
| 25 | 52214 | 51837 | 54307 | 51837 |
| 30 | 52214 | 51837 | 51837 | 51837 |
| 35 | 52214 | 51837 | 51837 | 51837 |
| 40 | 52214 | 51837 | 51837 | 51837 |
| 45 | 52214 | 51837 | 51837 | 51837 |

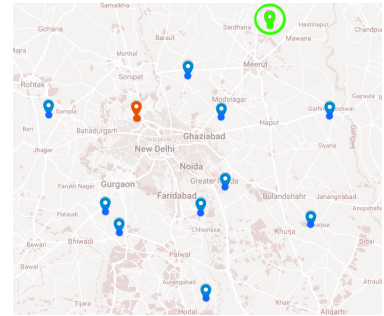

(a) Augmented set of 12 cities.

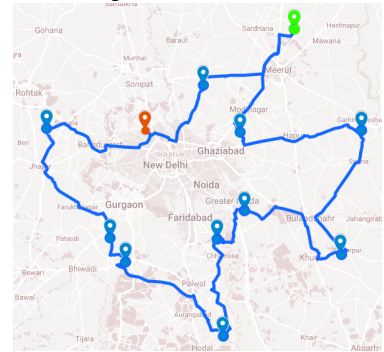

(b) Optimal Tour for the 12 cities.

Fig. 6: Maps depicting augmented set of 12 cities and its optimal tour.

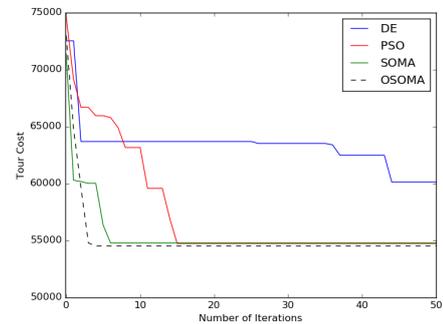

Fig. 7: Tour cost vs. Number of iterations for augmented set of 12 cities (For 50 iterations).

are applied on this set of 11 cities and the numerical results obtained have been presented in Table IV. OSOMA outperforms the other meta-heuristics considered as it reaches the optimal value in just 4 iterations while the other algorithms take more than 20 iterations. Fig. 4b shows the final optimal tour for this set of 11 cities. Fig. 5 depicts the tour cost of each algorithm as a function of the number of iterations.

*2) Case I:* In this case, a new city is added to the tour while the salesman is traveling. When the salesman was at city (shaded red), a new city (shaded green) is added to the initial set of 11 cities as can be seen in Fig. 6a. This results in an update in the cost matrix. Now, the salesman needs to take this change into account and re-evaluate the possible tours in order to find the new optimal tour. The opportunistic nature of OSOMA enables it to better adapt to this change as compared to SOMA, PSO and DE. This is evident from the numerical values in Table V. Fig. 6b presents the new optimal tour found for this augmented set of 12 cities. Fig. 7 depicts the tour cost of each algorithm as a function of the number of iterations. It can be easily seen that OSOMA attains the optimal value in only 3 iterations. This substantiates the claim that OSOMA achieves a higher accuracy solution at a faster speed as compared to the other algorithms.

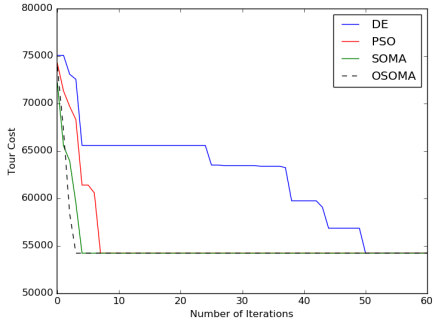

Fig. 8: Tour cost vs. Number of iterations for augmented set of 12 cities (for 60 iterations).

TABLE V: Tour Cost of DE, PSO, SOMA, OSOMA as a function of Number of Iterations (Case I)

| Iterations | SOMA | DE | PSO | OSOMA |
| --- | --- | --- | --- | --- |
| 0 | 71390 | 72531 | 74650 | 73011 |
| 5 | 56364 | 63697 | 65959 | 54526 |
| 10 | 54794 | 63697 | 63168 | 54526 |
| 15 | 54794 | 63697 | 54741 | 54526 |
| 20 | 54794 | 63697 | 54741 | 54526 |
| 25 | 54794 | 63697 | 54741 | 54526 |
| 30 | 54794 | 63523 | 54741 | 54526 |
| 35 | 54794 | 63523 | 54741 | 54526 |
| 40 | 54794 | 62484 | 54741 | 54526 |
| 45 | 54794 | 60133 | 54741 | 54526 |
| 50 | 54794 | 60133 | 54741 | 54526 |

TABLE VI: Tour Cost of DE, PSO, SOMA, OSOMA as a function of Number of Iterations (Case II)

| Iterations | SOMA | DE | PSO | OSOMA |
| --- | --- | --- | --- | --- |
| 0 | 72162 | 75078 | 74234 | 73913 |
| 5 | 54237 | 65590 | 61412 | 54237 |
| 10 | 54237 | 65590 | 54237 | 54237 |
| 15 | 54237 | 65590 | 54237 | 54237 |
| 20 | 54237 | 65590 | 54237 | 54237 |
| 25 | 54237 | 65590 | 54237 | 54237 |
| 30 | 54237 | 63519 | 54237 | 54237 |
| 35 | 54237 | 63462 | 54237 | 54237 |
| 40 | 54237 | 63396 | 54237 | 54237 |
| 45 | 54237 | 59754 | 54237 | 54237 |
| 50 | 54237 | 56867 | 54237 | 54237 |
| 55 | 54237 | 54237 | 54237 | 54237 |

*3) Case II:* This case considers the possibility of dynamic changes in the cost between any two cities, leading to an updation in cost matrix. Now, the salesman needs to take this change into account and re-evaluate the possible tours in order to find the new optimal tour. Table VI presents the performance of the considered algorithms with updated edge costs. The final optimal tour attained is the same as shown in Fig. 6, while the tour cost changes owing to the dynamic nature of the cost-matrix, as reflected in Table VI. Opportunistic nature of OSOMA allows it to adapt to this change in cost matrix efficiently. This is evident from Fig. 8 where OSOMA is able to reach the global minima in only 3 iterations. It should be noted that the optimal value attained has changed from 51,837 (in the Initial Case) to 54,237 (in this case).

## VI. CONCLUSION

SOMA is one of the most recent evolutionary optimization algorithms. A modified SOMA- Opportunistic SOMA has been presented which introduces a novel strategy to generate perturbations. This strategy enhances the capability of an individual by providing it with a higher number of choices than SOMA. This allows OSOMA to exploit the search space to a greater extent and thus, yield better solutions. It has been tested on several unconstrained benchmark test functions. The numerical results clearly show that OSOMA outperforms the other metaheuristic algorithms under consideration. The application of OSOMA on real-time DTSP further substantiates this claim. However, each algorithm has its own limitations and cannot be universally accepted to perform well on all optimization problems. Hence, in future work, it is intended to further improve the performance of this algorithm and apply it to various other practical problems.